\documentclass[12pt]{article}

\usepackage{scicite}
\usepackage{hyperref}
\usepackage{times}
\usepackage{graphicx}
\usepackage{caption}
\usepackage{xcolor}

\newenvironment{sciabstract}{ 
\begin{quote} \bf}
{\end{quote}}

\title{Empowering Machines to Think Like Chemists: Unveiling Molecular Structure-Polarity Relationships with Hierarchical Symbolic Regression}

\author
{Siyu Lou,$^{1,2\dagger}$ Chengchun Liu,$^{3\dagger}$ Yuntian Chen$^{2\ast}$ Fanyang Mo$^{3,4\ast}$\\
\\
\normalsize{$^{1}$Department of Computer Science and Engineering, Shanghai Jiao Tong University,}\\
\normalsize{Shanghai 200240, P.R.China}\\
\normalsize{$^{2}$Ningbo Institute of Digital Twin, Eastern Institute of Technology,}\\
\normalsize{Ningbo 315200, P.R.China}\\
\normalsize{$^{3}$School of Materials Science and Engineering, Peking University,}\\
\normalsize{Beijing 100871, P.R.China}\\
\normalsize{$^{4}$AI for Science (AI4S)-Preferred Program, Peking University Shenzhen Graduate School,}\\
\normalsize{Shenzhen 518500, P.R.China}
\\
\normalsize{$^\dagger$These authors contributed equally to this work.}
\\
\normalsize{$^\ast$To whom correspondence should be addressed;}\\
\normalsize{E-mail: ychen@eitech.edu.cn, fmo@pku.edu.cn.}
}

\date{}
\begin{document} 
\captionsetup[figure]{labelfont={bf},labelformat={default},labelsep=period,name={Fig.}}

\baselineskip21pt


\maketitle


\begin{sciabstract}
  Thin-layer chromatography (TLC) is a crucial technique in molecular polarity analysis. Despite its importance, the interpretability of predictive models for TLC, especially those driven by artificial intelligence, remains a challenge. Current approaches, utilizing either high-dimensional molecular fingerprints or domain-knowledge-driven feature engineering, often face a dilemma between expressiveness and interpretability. To bridge this gap, we introduce Unsupervised Hierarchical Symbolic Regression (UHiSR), combining hierarchical neural networks and symbolic regression. UHiSR automatically distills chemical-intuitive polarity indices, and discovers interpretable equations that link molecular structure to chromatographic behavior. 
\end{sciabstract}

% In setting up this template for *Science* papers, we've used both
% the \section* command and the \paragraph* command for topical
% divisions.  Which you use will of course depend on the type of paper
% you're writing.  Review Articles tend to have displayed headings, for
% which \section* is more appropriate; Research Articles, when they have
% formal topical divisions at all, tend to signal them with bold text
% that runs into the paragraph, for which \paragraph* is the right
% choice.  Either way, use the asterisk (*) modifier, as shown, to

\section*{Introduction}
Polarity reflects uniformity and symmetry of charge distributions in electrets \cite{sessler2005physical}. In particular, molecular polarity is crucial for understanding how molecules interact with each other, and it plays a fundamental role in the characterization of molecules \cite{Katritzky}. Thin-layer chromatography (TLC) has gained widespread acceptance\cite{Ciura,Poole} and provides critical insights into the behavior of organic molecules within varied solvent environments. Despite the established importance of TLC in determining the retardation factor ($R_f$), a crucial measure in polarity studies, the method is labor-intensive and involves large numbers of repetitive trials.

In recent years, significant progress has been made in the field of artificial intelligence-assisted chemical analysis, especially in the development of quantitative structure–activity relationships (QSAR) prediction models~\cite{zahrt2019,muratov2020qsar,wahl2021,rinehart2023} and quantitative structure–property relationships (QSPR) prediction models~\cite{yang2020holistic,yao2022,koscher2023}. There are some attempts have been made to construct prediction models for structure-retardation factors~\cite{KOMSTA200866,yousefinejad2015quantitative}, but their performance is constrained by the limited quantity and standardization of TLC data. To address this challenge, our prior work \cite{XU20223202,xu2022high} introduced an automatic high-throughput platform for TLC analysis, providing a more extensive and standardized dataset for model training. The enhanced dataset serves to improve the performance of the model, allowing for more accurate and robust predictions of structure-retardation factors in TLC experiments.

Although machine learning models, especially deep neural networks (DNNs) have demonstrated excellent performance, their ``black box'' nature raises concerns about interpretability and understanding of the underlying mechanisms. In this paper, we propose an innovative framework that integrates chemists’ experiential knowledge, aiming to ``unpack'' the ``black box'' of molecular structure-polarity relationships. 

Essentially, our approach seeks a formalized expression to reveal the intricate mathematical relationship between the $R_f$ value and the solute molecular structures as well as the eluent solvents in TLC analysis. Symbolic regression (SR) has recently emerged as a powerful tool for extracting the underlying mathematical equations from observed data~\cite{billard2003statistics,arnaldo2014multiple,quade2016prediction,brunton2016discovering,petersen2019deep,cranmer2020discovering,udrescu2020ai}. It has gained prominence for scientific discovery in diverse fields~\cite{sun2022symbolic,zeng2023deep,allen2022machine}. However, existing SR methods encounter performance limitations, particularly when dealing with datasets containing multiple variables. Therefore, when using SR methods to explain structure-polarity prediction models, a dilemma emerges: molecular fingerprints, such as MACCS key molecular fingerprints \cite{durant2002reoptimization}, despite their strong expressive power, pose great challenges for SR methods due to their high dimensionality; on the other hand, utilizing feature engineering and domain knowledge with a limited set of input variables results in discovered formulas lacking expressiveness. 

\begin{figure}
\centering
  \includegraphics[width=\textwidth]{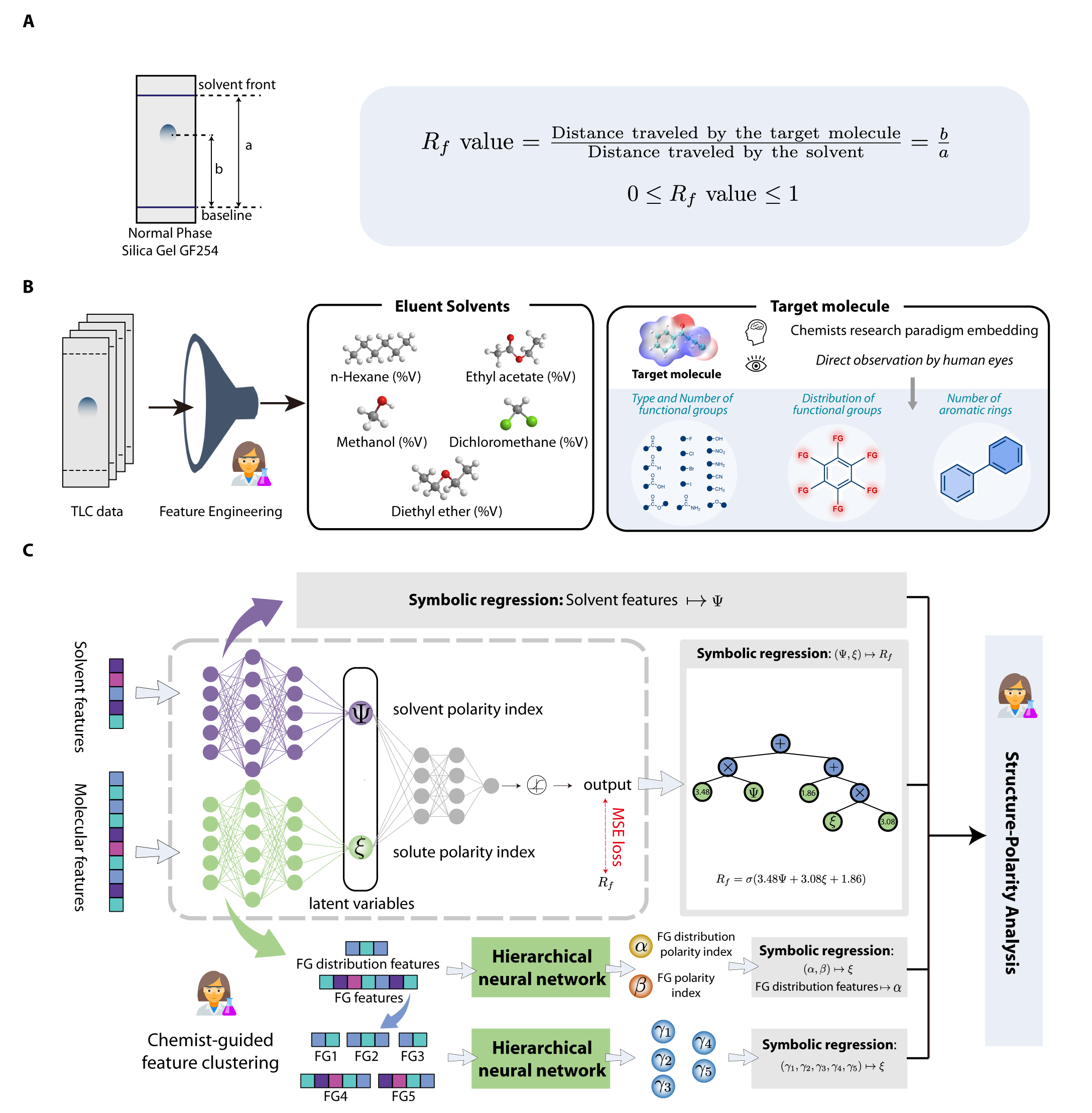}
  \vspace{-1em}
  \caption{\textbf{Overview of Unsupervised Hierarchical Symbolic Regression (UHiSR).} \textbf{(A)} Illustration of TLC experiment and the calculation of the retardation factor ($R_f)$. \textbf{(B)} Feature engineering, involving five solvent features based on volume percentages and the decomposition of target molecules into functional groups (FG). The molecular structure is treated as a composite formed by stacking various functional group modules. \textbf{(C)} UHiSR framework with three main stages: chemist-guided feature clustering, hierarchical neural network for latent variable extraction (\emph{e.g.}, solute polarity index), and symbolic regression for discovering explicit equations between the target value and latent variables. }
  \label{Fig.1}
\end{figure}

To address this dilemma between expressiveness and interpretability, we present unsupervised hierarchical symbolic regression (UHiSR) (see~Fig.~\ref{Fig.1}), a new SR approach guided by hierarchical neural network. UHiSR introduces novel polarity indices, \emph{e.g.}, solvent polarity index and solute polarity index, which are learned through a semantic hierarchical model. Guided by domain knowledge, this model comprises multiple sub-models that govern the mapping relationships between a set of input variables, \emph{e.g.}, solvent compositions, and a specific polarity index, \emph{e.g.}, solvent polarity index, as well as the mapping between polarity indices and the output variable $R_f$. It is noteworthy that this process aligns with the chemists' thinking processes. Human brain struggles with high-dimensional information, opting not to directly comprehend intricate mappings but rather to decompose and analyze through individual sub-models. Then, we can apply the SR method on the output $R_f$ and the polarity indices, effectively reducing the dimension of the input variables of SR algorithm.

In addition, experienced organic chemists often do not use the molecular fingerprints to evaluate what solvent system should be used for TLC experiments for a given target molecule. Instead, organic chemists evaluate polarity primarily based on the nature, number, and molecular structure of functional groups in solute compounds \cite{sherma2003handbook}. Therefore, in this study, we seek a simpler and more chemically intuitive perspective, focusing on the basic characteristics of complex molecules, moving from computer engineering to a human-scientist-like thinking mode. 

In conclusion, our work introduces several novel chemical-intuitive polarity indices, including the solvent polarity index and solute polarity index, in TLC experiments. These indices provide quantitative insights into the interaction dynamics among molecules in both solid-liquid and liquid-liquid systems. Furthermore, we present a concise-yet-accurate formula that links the $R_f$ value with these polarity indices. The UHiSR framework, driven by both data and domain knowledge, not only uncovers these meaningful polarity indices directly from experimental data, but also constructs an equation that governs the underlying mechanisms. It empowers the \textit{machine to think like chemists}, and builds a bridge between AI models and interpretable insights in molecular structure-polarity predictive modeling.

\section*{Results}
\subsection*{Data driven interpretable polarity indices}

It is essential for chemists to accurately characterize the overall polarity of molecules. For this purpose, there are many known descriptors such as the molecular polarity index (MPI), which is based on theoretical calculations. However, they often come with issues like high computational cost or only providing global information. Another way to estimate the molecule's polarity is based on TLC experiments, where chemists typically keep solutes or solvents fixed, evaluate different solvent compounds or solute compounds, and utilize the $R_f$ value to infer the corresponding polarities. 

Based on the TLC dataset, we used a data-driven method to extract two polarity indices, namely $\Psi$ and $\xi$, acting as empirical descriptors providing information about the polarity of the solvent system and solute molecule, respectively. In contrast to traditional physicochemical descriptors, our proposed polarity indices offer enhanced interpretability and the ability to capture detailed structural information.

\paragraph*{Chemical-intuitive feature engineering.}Rather than employing molecular fingerprints or conventional physicochemical descriptors like common machine learning does, our approach deliberately labels functional groups, just like human chemists typically do. In particular, this study encompasses an array of solute molecules, including those containing carbonyl, hydroxyl, amino, halogen, nitro, and cyano functionalities. This enables a tailored analysis of how these functional groups influence the compound's $R_f$ value across various solvents in terms of type, number, and allocation. For notational simplicity, we utilize the abbreviations from Table~\ref{tab:feature} for the features employed in our study. This table also provides the descriptions and statistics of these features. Our method merely uses $19$ distinct features for describing solute molecules, and thereby demonstrates an effective reduction of dimensionality compared with MACCS key molecular fingerprints ($167$ dimensions). Furthermore, in contrast to traditional physicochemical descriptors, which are commonly low dimensional, our features offer a more localized and interpretable representation of the target molecule.

\begin{figure}
 \centering
  \includegraphics[width=1\linewidth]{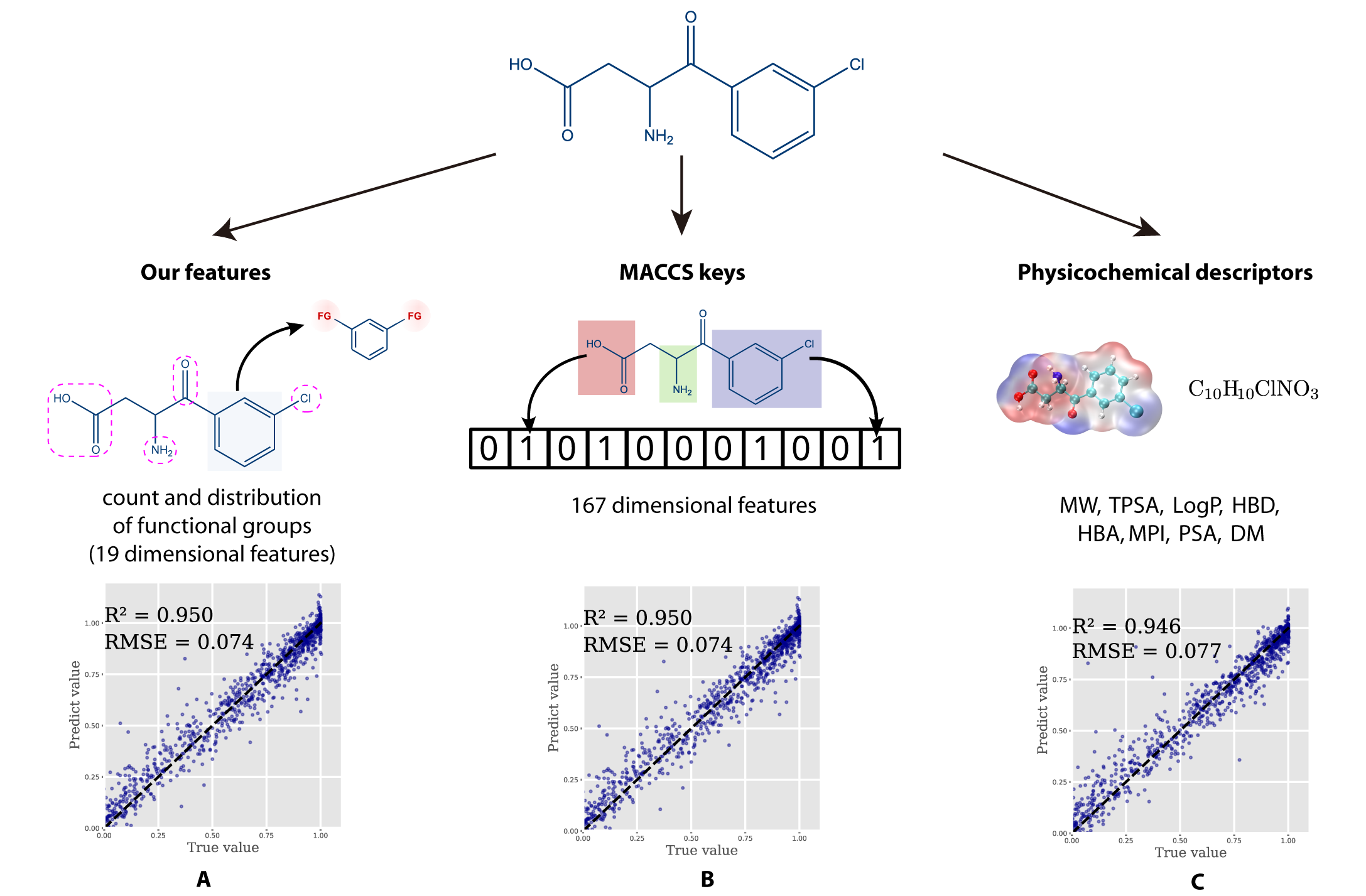}
  \caption{\textbf{Comparative analysis of different molecular feature sets}: the fitting accuracy is evaluated using the XGBoost model across three different feature groups. These feature groups comprise: (A) Features introduced in this paper, detailed in Table~\ref{tab:feature}; (B) MACCS keys; (C) physicochemical descriptors.}
  \label{fig:feature}
\end{figure}

To evaluate the effectiveness of the selected features, we conducted a comparative analysis of their representation power (see Materials and Method for more details). As shown in Fig.~\ref{fig:feature}, our proposed features demonstrate comparable fitting performance to MACCS, and slightly surpass conventional physicochemical descriptors. The results emphasize the substantial representational power of our features.

\begin{table}
\centering
\caption{\textbf{Description and statistic analysis of the features used in this study.} The abbreviations, descriptions, mean values along with the standard deviations (Std.) are presented. Additionally, Pearson's correlation coefficients ($r$) are computed to quantify the linear relationship between each feature and the observed $R_f$ value.}
\begin{tabular}{@{}llccc@{}}
\hline
Feature abbr. & Feature description & Mean(Std.) & Pearson's $r$\\
\hline
\multicolumn{4}{c}{Eluent solvent} \\
\hline
Hex & n-Hexane concentration & $0.36(0.40)$ & $-0.54$ \\
EA & Ethyl acetate concentration & $0.15(0.28)$ & $0.29$\\
DCM & Dichloromethane concentration & $0.40(0.48)$ & $0.21$\\
MeOH & Methanol concentration&$0.014(0.025)$ & $0.31$\\
Et$_2$O & Diethyl ether concentration &$0.071(0.22)$& $0.13$\\
\hline
\multicolumn{4}{c}{Solute molecule}\\
\hline
NBen & Count of benzene  & $1.12(0.57)$& $0.07$\\
MSD & Molecular skeleton descriptor & $3.79(2.32)$ & $0.07$\\
DM & Dipole moment  & $1.34(0.74)$ & $-0.02$\\
CtPhenol & Count of phenolic hydroxyl group & $0.15(0.40)$ & $-0.15$ \\
CtOH & Count of alcoholic hydroxyl group &$0.04(0.21)$ & $-0.13$\\
CtAldehyde & Count of aldehyde group  & $0.19(0.40)$ & $-0.007$ \\
CtCO$_2$H & Count of carboxylic acid group & $0.03(0.18)$ & $-0.17$\\
CtRCO$_2$R & Count of ester group & $0.10(0.32)$ & $0.03$ \\
CtR$_2$C=O & Count of ketone group & $0.35(0.54)$ & $-0.06$\\
CtROR & Count of ether group &$0.17(0.47)$ & $-0.04$\\
CtCN & Count of cyano group &$0.06(0.25)$ & $-0.01$\\
CtNH$_2$ & Count of amine group & $0.08(0.28)$ & $-0.20$\\
CtNO$_2$ &  Count of nitro group & $0.06(0.25)$ & $0.003$\\
CtAmide & Count of amide group & $0.02(0.14)$ & $-0.14$\\
CtMe & Count of methyl group & $0.20(0.54)$ & $0.10$\\
CtF & Count of fluorine  & $0.25(0.81)$ & $0.07$ \\
CtCl & Count of chlorine  & $0.11(0.39)$ & $0.13$\\
CtBr & Count of bromine & $0.15(0.39)$ & $0.15$\\
CtI & Count of iodine  & $0.12(0.33)$ & $0.22$\\
\hline
\end{tabular}
\label{tab:feature}
\end{table}

\begin{figure}
  \includegraphics[width=1.00\textwidth]{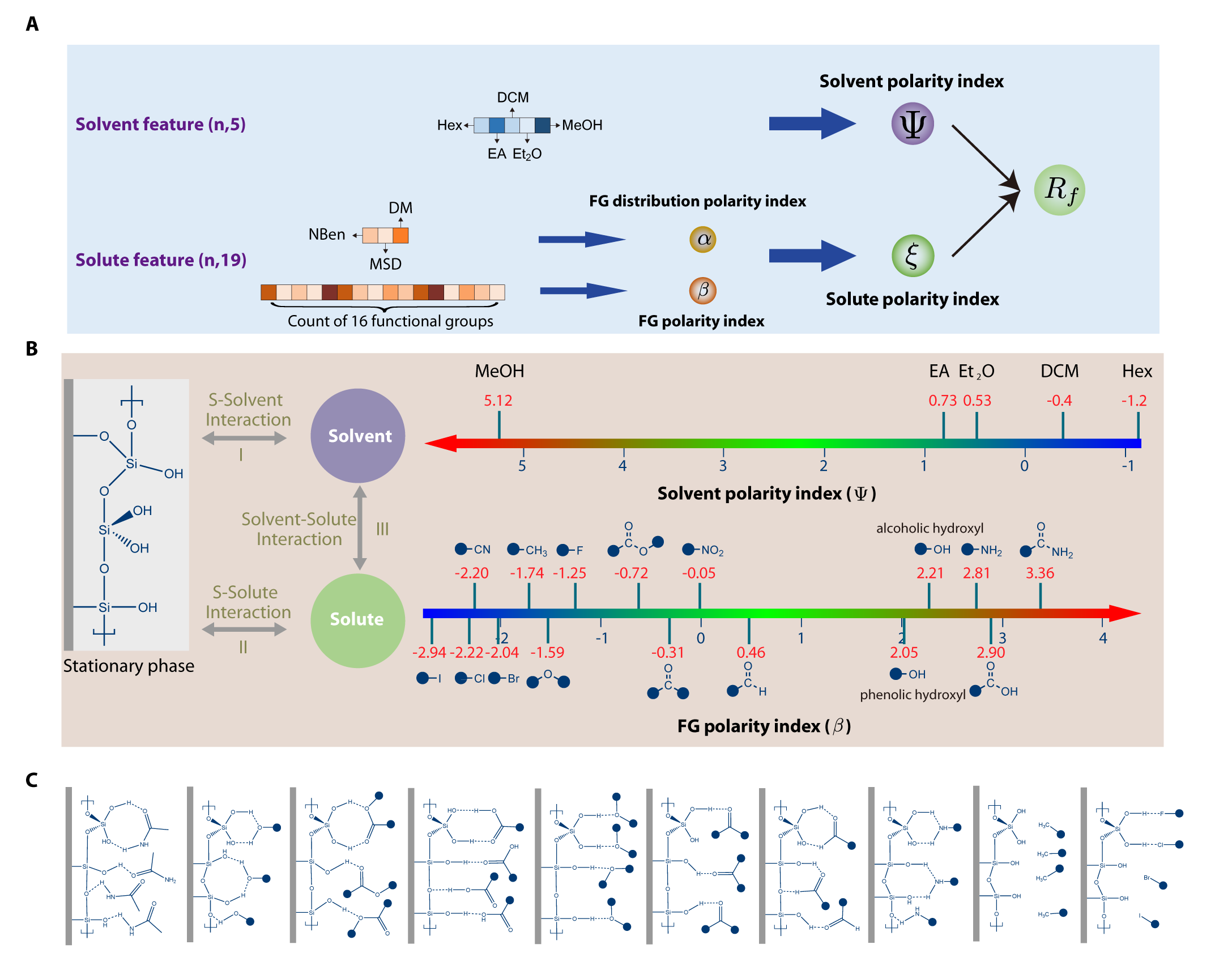} 
  \caption{\textbf{Illustration of the polarity indices and their impact on chromatographic behavior.} (\textbf{A}) The input features corresponds to different polarity indices. Here, FG stands for functional group. (\textbf{B}) TLC experiment can be understood as a process where the stationary phase (silica gel) and the mobile phase (solvent) compete for the solute molecules. This competition's outcome is reflected in the $R_f$ value. The factors influencing the $R_f$ value can be categorized into three types: interaction between the stationary phase and the solvent (I), interaction between the stationary phase and the solute (II), and interaction between the solvent and the solute (III). The two polarity indices ($\Psi$ and $\xi)$ characterize how solvent and solute impact the chromatographic behavior separately. (\textbf{C}) Illustration of interactions between the stationary phase and different functional groups.}
  \label{fig:solvent_analysis_1}
\end{figure}

\paragraph*{Polarity indices.} In this study, the stationary phase on the TLC plate is silica gel. The determined $R_f$ values epitomize a dynamic equilibrium. It reflects the competitive interplay between the highly polar stationary phase and the solute molecules, which are mediated through interactive forces by changing the polarities of the mobile phase during the experiment. This interplay exhibits two primary patterns: the solvent facilitates the upward transit of solute via capillary action; concurrently, solvent molecules intervene by dislodging solute molecules from the stationary phase's surface, attenuating their interactions (as illustrated in Fig.~\ref{fig:solvent_analysis_1}). 

The two polarity indices distinctly capture the interaction dynamics between the solvent compound and the silica gel, as well as between the solute molecule and the silica gel. Importantly, our approach enables a quantitative assessment of the aforementioned interactions. In the following paragraphs, we will separately analyze these two indices.

\paragraph*{Solute polarity index.} Considering that the dataset contains $387$ organic compounds, previous research normally involves high dimensional features and physicochemical descriptors to encode the solute molecule into numerical values. However, the former are often challenging for human understanding, while the latter lack structural information. 

To carry out this study, we carefully selected $16$ representative functional groups and used their counts as features for the target molecules. Additionally, we characterized the distribution of these functional groups within the molecule by using three features: count of benzene (NBen), molecular skeleton descriptor (MSD), and dipole moment (DM). Although we were able to reduce the feature numbers to $19$, establishing a explainable connection between the polarity of the molecule and these input features remains a major challenge. 

We distinguished two types of solute features: functional group (FG) counts and FG distribution, as they may have distinct effects on the polarity of a given molecule (Fig.~\ref{fig:solvent_analysis_1}A). For instance, a molecule with a higher count of polarity-functional groups but with a symmetric structure may exhibit lower polarity than molecules with fewer polarity-functional groups. Therefore, we further extracted two indices, namely FG distribution polarity index $\alpha$ and FG polarity index $\beta$, to isolate these effects. 

Despite the polarity of functional groups being of great interest, there is a lack of quantitative and scalable methods to compare the polarities of different functional groups; in current practice, chemists often rely mainly on qualitative analysis and their experience. Our FG polarity index provides the first empirical quantification of the impact of different functional groups on molecular polarity. Considering the $16$ functional groups involved in this study, the input features for the sub-model used to obtain the FG polarity index $\beta$ consist of $16$-dimensional vectors, with each dimension representing the number of the respective functional group. To quantify the polarity of a specific functional group FG$_i$, we configured the input feature vector such that the $i$-th position was set to $1$ and the others to $0$. As illustrated in Fig.~\ref{fig:solvent_analysis_1}B, the sub-model's outputs reveal a distinct polarity order among the functional groups, \emph{i.e.}, amides ($3.36$) $>$ carboxylic acid ($2.90$) $>$ amine ($2.81$) $>$ alcoholic hydroxyl ($2.21$) $>$ phenolic hydroxyl ($2.05$) $>$ aldehyde ($0.46$) $>$ nitro ($-0.05$) $>$ ketone ($-0.31$) $>$ ester ($-0.72$) $>$ fluorine ($-1.25$)$>$ ether ($-1.59$) $>$ methyl ($-1.74$) $>$ bromine ($-2.04$) $>$ cyano ($-2.20$) $>$ chlorine ($-2.22$) $>$ iodine ($-2.94$).  

\paragraph*{Solvent polarity index.}Another important interaction is between the solvent compound and the silica gel. The solvent polarity index $\Psi$ provides an analytical lens through which we are able to assess the influence of disparate solvents on the $R_f$ values. 

Methanol (MeOH) possesses one hydrogen bond acceptor (HBA) and one hydrogen bond donor (HBD), capable of forming stable six- and seven-membered chelates with the silica gel. It also boasts an HBA that can engage in intermolecular hydrogen bonding with the siloxane (Si-O-Si) bridges of the silica gel. Ethyl acetate (EA) harbors two HBAs that can establish an eight-membered ring complex with the silica gel. The oxygen atoms of the ester moiety can partake in two distinct types of intermolecular hydrogen bonds with the gel. Diethyl ether (Et$_2$O) contains one HBA and is thus limited to a singular mode of intermolecular hydrogen bonding with the silica gel. Dichloromethane (DCM) is characterized by two chlorine atoms, which can form a relatively less stable eight-membered ring structure with the gel, and induce weak intermolecular hydrogen bonding. The interaction between n-hexane (Hex) molecules and the silica gel is considerably feeble, with negligible capacity to displace solute molecules. Consequently, under the paradigm of ``like dissolves like'', we infer a polarity hierarchy as follows: 
\[
{\rm MeOH} > {\rm EA} > {\rm Et}_2{\rm O} > {\rm DCM} > {\rm Hex}.
\]

The input features of the sub-model used to obtain $\Psi$ consist of five-dimensional vectors, with each dimension representing the volume percentage of a solvent compound. Consequently, variations in the compound ratio within the solvent system directly influence the value of $\Psi$, thereby changing the polarity of the solvent.  Moreover, to isolate the effect of each compound, we can assign the value $1$ to the for the volume percentage that corresponds to the target solvent compound, while setting the others to $0$. This approach yields the following values (see Fig.~\ref{fig:solvent_analysis_1}B): 
\[\Psi_{{\rm Hex}} = -1.238, \quad\Psi_{{\rm DCM}} =-0.392, \quad \Psi_{{\rm Et}_2{\rm O}} = 0.534, \quad \Psi_{{\rm EA}} = 0.733,\quad\Psi_{{\rm MeOH}} = 5.124. 
\]

As shown in Table~\ref{tab:solvent}, we also conducted a comparative analysis for our empirical solvent descriptor $\Psi$ and experimental miscibility indices such as LogP and aqueous solubility\cite{smallwood2012handbook}. The observed numerical consistency and trends align with our preliminary postulates (see Table~\ref{tab:solvent}).

\begin{table}
\caption{Solvent properties and solvent polarity index $\Psi$.}
\centering
\footnotesize % This will make the font size of the entire table smaller
\begin{tabular}{@{}lcccccc@{}}
\hline
Solvent & HBD & HBA & Solubility in water (g/100g) \cite{smallwood2012handbook} & LogP &  $\Psi$ \\
\hline
Hex & 0 & 0 & 0.01 & 3.42 & -1.238\\
DCM & 0 & 0& 1.32 & 1.01 &-0.392\\
Et$_2$O & 0 & 1 & 6.9  & 0.76  & 0.534\\
EA & 0 & 2 & 7.7 & 0.29  &0.733\\
MeOH & 1 & 1 & Miscible & -0.27 &5.124\\
\hline
\end{tabular}
\label{tab:solvent}
\end{table}

\subsection*{The $R_f$ governing equation}
To comprehensively understand the relationship between the $R_f$ value and the two polarity indices, we used the symbolic regression to formulate empirical mathematical expressions connecting the $R_f$ value and the two polarity indices (Fig.~\ref{fig:sigmoid}A). The observed $R_f$ values were obtained directly from our automatic high-throughput platform \cite{xu2022high}, while the computed $R_f$ values were calculated from the $R_f$ governing equation (see Equation (1) below). The polarity indices can also be determined through a hierarchical equation system, based on the values of the input variables (Fig.~\ref{fig:sigmoid}A), a discussion of which will follow.

To ensure that the computed $R_f$ values lie within the range from $0$ to $1$, a sigmoidal function was employed (see Fig.~\ref{fig:sigmoid}B). By using symbolic regression, one obtains several candidates for the $R_f$ governing equation (see Table S1 in the Supplementary Materials). Since a simple formula is desired, and considering the prescribed fitting accuracy (RMSE $=0.091$, $R^2=0.93$), we selected 

\begin{equation}
    R_f= \sigma\left(3.48 \Psi + 3.08 \xi + 1.86\right),\quad \sigma(x) = \frac{1}{1+e^{-x}}.
\label{eq:equation1}
\end{equation}

In order to separately analyze the individual influence of solvent and solute on the $R_f$ value, we further decomposed the formula for $R_f$ in Equation (1) into $h(\Psi)$ and $g(\xi)$. For instance, as illustrated in Fig.~\ref{fig:sigmoid}C, when considering a specific solute molecule (\emph{i.e.}, fixing a value for $\xi$), we can reformulate Equation (1) as $R_f = h(\Psi) = \sigma(3.48\Psi+C_1)$, where $C_1$ depends on the chosen solute molecule. The pattern observed in Fig.~\ref{fig:sigmoid}C shows that $R_f$ increases with $\Psi$. Since a larger value of $\Psi$ indicates greater polarity, this finding emphasizes that the $\Psi$ correlates to the polarity of the solvent. Furthermore, we observe that the shape of $h(\Psi)$'s graph varies significantly for different solute molecules. In particular, Fig.~\ref{fig:sigmoid}C indicates that, within the context of our dataset, changing the solvent's composition only has a subtle effect on $R_f$ for solute molecules with extreme polarities. Similarly, when the solvent is fixed, Equation (1) can be reformulated as $R_f=g(\xi) = \sigma(3.08\xi + C_2)$, where $C_2$ depends on the chosen value for $\Psi$. By analysis similar to the previous case (fixed solute), the $R_f$ value increases with $\xi$. But now, a larger value of $\xi$ indicates a lower polarity. Moreover, we observe that the shape of $g(\xi)$'s graph is of the same S-shaped type across different solvent systems (see Fig.~\ref{fig:sigmoid}C).

\subsection*{Quantification of solvent polarity}

In addition to characterizing the polarity of the eluent solvent, the UHiSR framework offers an explainable solution for understanding how different compounds influence the solvent polarity index $\Psi$. In this study, we consider five distinct solvent compounds: Hex, EA, DCM, MeOH and Et$_2$O. To achieve the designed solvent polarity, chemists usually adjust the ratios of these compounds in the mobile phase system. By directly applying the SR algorithm, we obtained the empirical mathematical equation between solvent polarity index $\Psi$ and its corresponding input variables, \emph{i.e.}, the volume ratio of Hex, DCM, Et$_2$O, EA and MeOH in the mobile phase system. The following formula achieved high $R^2$ of $0.983$ and a low RMSE of $0.094$ with a simple form,
\begin{equation}
    \Psi=- {\rm Hex} + 1.59 {\rm EA} - 0.411 {\rm DCM} + 11.1 {\rm MeOH} + {\rm Et}_2{\rm O}^{2} + 0.142.
\end{equation}
Specifically, there are three mobile phase systems involved in the dataset, which are Hex/EA, Hex/Et$_2$O and MeOH/DCM. Equation (2) quantifies the impact of the two compounds on solvent polarity within each system. For instance, in the presence of Hex and EA, the solvent polarity index is calculated as $\Psi = -{\rm Hex} + 1.59{\rm EA}$. As the volume of EA increases, the solvent polarity exhibits a corresponding increase. Moreover, the constant term of 11.1 with MeOH indicates that even a slight increase/decrease in the volume of MeOH will lead to a significant increase/decrease in solvent polarity.

\subsection*{Quantification of solute polarity}
As mentioned earlier, we introduce two additional functional group-related indices, FG distribution polarity index $\alpha$ and FG polarity index $\beta$, to enhance our understanding of the relationship between the molecular polarity and the molecular structure. With the help of symbolic regression, we obtained an explicit mathematical relationship between $\xi$ and $(\alpha, \beta)$ as follows:
\begin{equation}
    \xi = c_1\beta + c_2\alpha e^{\alpha} + c_3e^{\alpha} + c_4\alpha + c_5 e^{\alpha}\beta + c_6,
\end{equation}
where $c_1=-0.257, c_2=-0.464, c_3=1.093, c_4=0.0246, c_5=0.464, c_6=-0.058$. Breaking down Equation (3), the molecular polarity can be understood as a combination of the polarity of the functional groups within the molecule ($f_1(\alpha)=c_2\alpha e^{\alpha} + c_3e^{\alpha} + c_4\alpha$), the distribution of these functional groups ($f_2(\beta)=c_1\beta$) and the interaction between these two factors ($f_3(\alpha, \beta)=c_5 e^{\alpha}\beta + c_6$).  

\begin{figure}
  \includegraphics[width=1.00\textwidth]{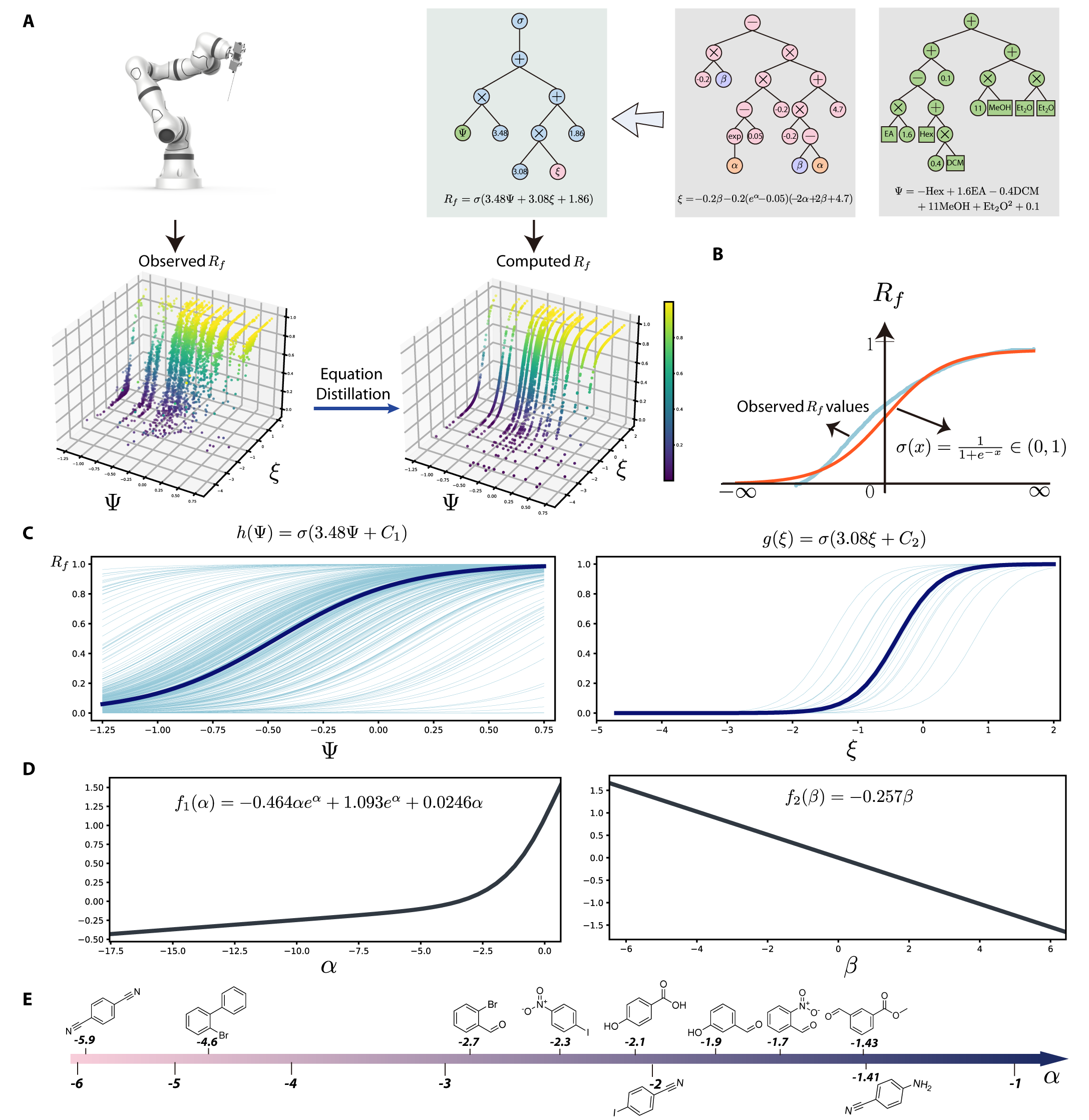}
  \caption{\textbf{Visualization of the latent variables and the decomposition of the retrieved formula.} (\textbf{A})Visualization of the observed and calculated $R_f$ values with two polarity indices $\Psi$ and $\xi$. (\textbf{B}) Fitting observed Rf values with a Sigmoid function. (\textbf{C}) Decomposition of Equation (1) into $h(\Psi)$ and $g(\xi)$. (\textbf{D})   Decomposition of Equation (3) into $f_1(\alpha)$ and $f_2(\beta)$. (\textbf{E}) FG distribution polarity index $\alpha$ of 10 example compounds.} 
  \label{fig:sigmoid}
\end{figure}

Regarding the FG polarity index $\beta$, which categorizes the polarity of different functional groups, we observe a distinct trend: an increase in $\beta$ corresponds to a decrease in the solute polarity index $\xi$, consequently leading to a smaller $R_f$ value (Fig.~\ref{fig:sigmoid}D). This signifies that a higher $\beta$ value is indicative of greater polarity. As for the FG distribution polarity index $\alpha$, we observe an exponential growth of $\xi$ when $\alpha$ increases. However, $\alpha$ is a composite element that includes the functional group distribution within- and the symmetry of the target molecule. In Fig.~\ref{fig:sigmoid}E, we provide $10$ example molecules with distinct $\alpha$ values.

\subsection*{Hierarchical equation system} UHiSR has the capability to reveal the relationship between the FG distribution polarity index $\alpha$ and its corresponding input variables, as well as between the FG polarity index $\beta$ and the counts of $16$ functional groups. Consequently, we can establish a hierarchical equation system that fully elucidates the mathematical relationship between $R_f$ and all input variables, as presented in Table~\ref{tab:equation_system}. 

\begin{table}
\centering
\caption{Optimal hierarchical equation systems corresponding to $R_f$ and input variables.}
\begin{tabular}{@{}lccc@{}}
\hline
Equation & $R^2$ &RMSE \\
\hline
$R_f = \sigma\left(3.48 \Psi + 3.08 \xi + 1.86\right),\quad \sigma(x) = 1/(1+e^{-x})$ & $0.930$ &$0.091$  \\[0.2cm]
$\Psi=-{\rm Hex} + 1.59 {\rm EA} - 0.411 {\rm DCM} + 11.1 {\rm MeOH} + {\rm Et}_2{\rm O}^{2} + 0.142$ & $0.983$ & $0.094$ \\[0.2cm]
$\xi = - 0.232 \beta - 0.232 \left(e^{\alpha} - 0.0531\right) \left(- 2 \alpha + 2 \beta + 4.71\right)$ & $0.841$ & $0.270$ \\[0.2cm]
\begin{minipage}{0.8\linewidth}\hspace{2em}$ \alpha =- 2 {\rm NBen}\left({\rm DM} + 0.412\right) - \frac{1.33 {\rm NBen}\left({\rm DM} + 0.412\right)}{{\rm MSD} - 0.0467} - 0.743 $ \end{minipage} & $0.898$ & $0.63$ \\[0.2cm]
\begin{minipage}{0.8\linewidth}\hspace{2em}$  \beta =-0.218 \gamma_1/\gamma_2^{6} + 0.413 \gamma_2 + 0.435 \gamma_4 - 0.435 \gamma_5 + 0.0223 \left(\gamma_3 + \gamma_4\right)^{2} $\\$
    ~\quad~\quad~\quad~+ 0.493 $ \end{minipage}  & $0.810$ & $0.934$ \\[0.2cm]
\begin{minipage}{0.8\linewidth}\hspace{2em}$\gamma_1 = - 3.09 {\rm CtAmides} - 3.91 {\rm CtCO}_2{\rm H} + 1.87 $ \end{minipage}  & $0.997$ & $0.152$ \\[0.2cm]
\begin{minipage}{0.8\linewidth}\hspace{2em}$\gamma_2  = - {\rm CtRNH}_2 + 2 {\rm CtOH} - 1.76 {\rm CtPhenol}  \left(1.76 - {\rm CtRNH}_2\right) $\\$
   ~\quad~\quad~\quad~~+ \left(- {\rm CtRNH}_2^{2} + {\rm CtOH} - {\rm CtPhenol} + 0.912\right)^{2} $ \end{minipage}  & $0.999$ & $0.204$ \\[0.2cm]
\begin{minipage}{0.8\linewidth}\hspace{2em}$\gamma_3 ={\rm CtNO}_2 \left(- {\rm CtRCO}_2{\rm R}^{2} - 3.65\right) + 0.762 $ \end{minipage}  & $0.961$ & $0.413$ \\[0.2cm]
\begin{minipage}{0.8\linewidth}\hspace{2em}$\gamma_4 = {\rm CtAldehyde}^{3} - {\rm CtAldehyde}^{2} \left({\rm CtR}_2{\rm C}\!\!=\!\!{\rm O} - {\rm CtF}\right)^{2}$\\$
    ~\quad~\quad~\quad~~ - 3 {\rm CtAldehyde} + {\rm CtR}_2{\rm C}\!\!=\!\!{\rm O}^{2} + 2 {\rm CtR}_2{\rm C}\!\!=\!\!{\rm O} + 4 {\rm CtCN}  $\\$
    ~\quad~\quad~\quad~~- 2 {\rm CtF} + e^{{\rm CtF} - \left({\rm CtROR} - 2 {\rm CtF}\right)^{2}}- 0.746$
    \end{minipage}  & $0.961$ & $0.446$ \\[0.2cm]
\begin{minipage}{0.8\linewidth}\hspace{2em}$ \gamma_5= \left({\rm CtCl} + 2 {\rm CtI}\right)^{2} + \left(\frac{{\rm CtBr}}{{\rm CtBr} - 0.305} + {\rm CtMethyl}\right)^{2}$ \end{minipage}  & $0.931$ & $0.805$ \\[0.2cm]
\hline
\end{tabular}
\label{tab:equation_system}
\end{table}

\section*{Discussion}
Recent methods to identify quantitative structure-polarity relationships usually include various machine learning models and learning informative feature representations~\cite{yang2020holistic,xu2022high}. Despite achieving superior prediction accuracy, these models or feature representations often lack interpretability, hindering a deeper understanding the underlying relationships within the dataset. One strategy to enhance interpretability is to articulate explicit formulas between the output and input variables. Symbolic regression emerges as an effective approach for unveiling complex relationships within datasets. However, feasibility and interpretability are challenged when dealing with a large number of variables, as the performance significantly decreases while the complexity of the discovered formulas increases drastically.

In this work, we introduce the integration of hierarchical neural networks and symbolic regression, aiming to enhance the interpretability of the model. We propose, for the first time, two important polarity indices, the solvent polarity index $\Psi$ and the solute polarity index $\xi$. These two polarity indices not only provide interpretable insights, but also enable direct $R_f$ prediction through a formula-driven approach. Compared to high-dimensional feature representations, as we only use few indices (two) that are strongly tied to the given dataset, our approach offers increased ease of understanding and better alignment with the underlying task and the dataset. Additionally, the predictive accuracy is comparable to that of currently used DNN models. Besides, the data-driven discovery of FG distribution polarity index and FG polarity index facilitates a systematic exploration of our $19$-dimensional solute feature space, with each index focusing on different aspects (counts and distribution of functional groups).

The integration of symbolic regression offers an equation-based model, which has several advantages compared with traditional machine learning models. First, equation-based models have extremely low computational cost, making it an ideal choice in resource-constrained environments. Second, the equation-based model usually has only few parameters, especially compared to DNN models, increasing its transferablility and robustness. Due to these advantages, there are potential applications of equation-based models in edge computing. Moreover, unlike traditional neural network models that are deployed in centralized systems and require substantial data transfer for each new scenario, equation-based models offer much cheaper adaptability. This opens up new possibilities for applications in areas such as automated experimental platforms.  

As final remark, we have focused only on examples from the chemistry community; however, building interpretable frameworks, that facilitate human understanding of the employed models, can be of general importance for science applications. 

\section*{Materials and Methods}
\subsection*{TLC Dataset}
Several variables contribute to the variation in $R_f$ values, including factors such as the nature of the stationary phase, temperature, humidity. Traditional methods of obtaining TLC data manually suffer from a lack of standardization, making scalability challenging. To address this issue, our prior work \cite{xu2022high} introduced a robotic platform designed for high-throughput collection of TLC data (Fig.~\ref{fig:sigmoid}A), which not only ensures efficiency in data acquisition but also establishes a standardized and scalable methodology.

The TLC dataset comprises in total $4,944$ measurements of the $R_f$ values, involving $387$ organic compounds and three mobile phase systems. These mobile phase systems are \textit{n}-Hexane/Ethyl acetate, Diethyl ether/\textit{n}-Hexane and Methanol/Dichloromethane systems, with a total of $17$ different solvent compositions.

\begin{figure}
\centering
  \includegraphics[width=\textwidth]{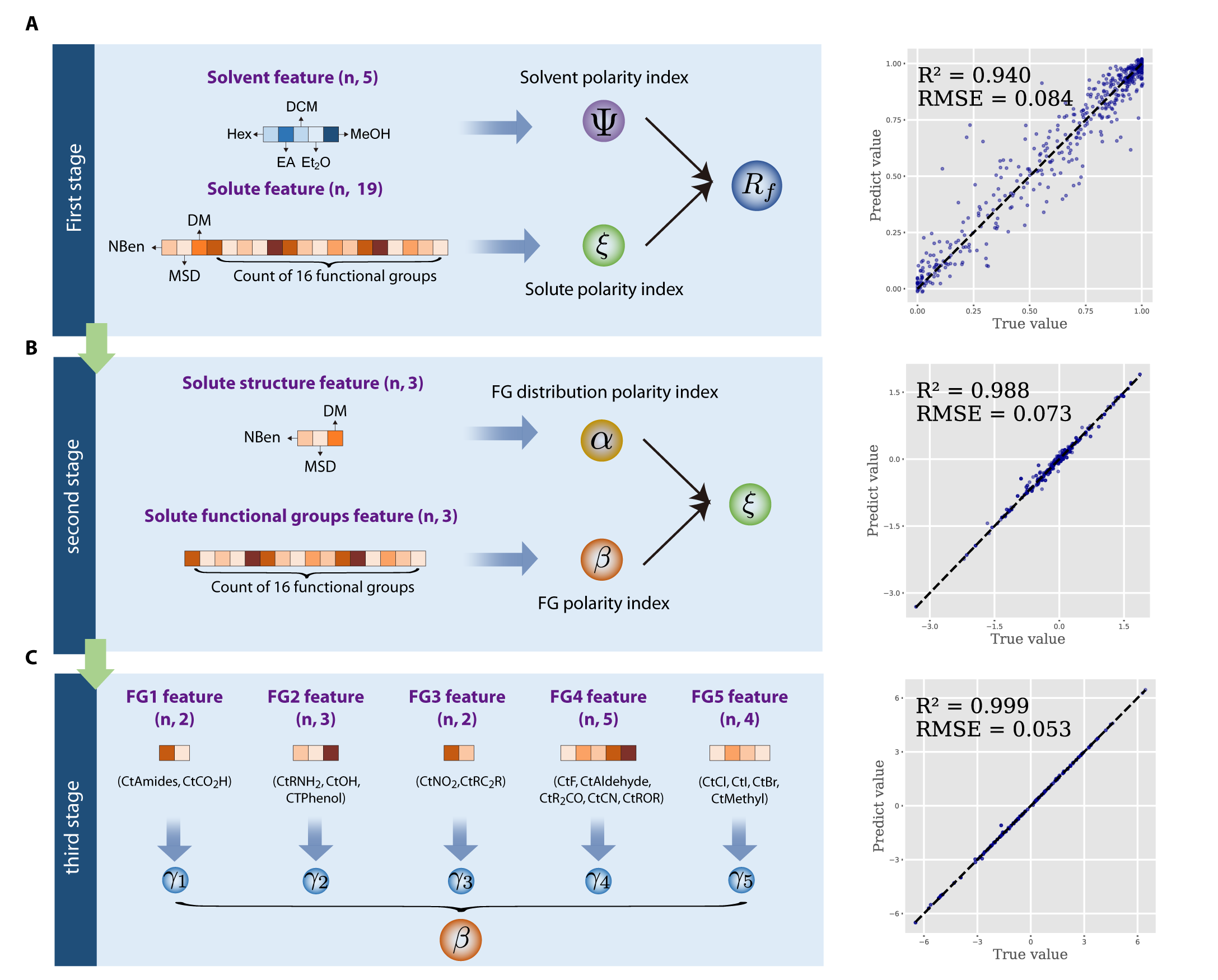}
  \caption{\textbf{Hierarchical structure of learning latent variables.} Three stages are organized from top to bottom. (\textbf{A}) At the first stage, two latent variables, $\Psi$ and $\xi$, are generated to encapsulate the overall polarity of the solvent and the solute, respectively. (\textbf{B}) At the second stage, two additional latent variables, $\alpha$ and $\beta$, are learned to assess the solute molecule's polarity from two distinct perspectives. $\alpha$ is linked to the distribution of functional groups, while $\beta$ pertains to the quantity of individual functional groups. (\textbf{C}) The third stage characterizes five latent variables, each representing the impact of specific groups of functional groups.}
  \label{fig:learn_variable}
\end{figure}

\subsection*{Framework of UHiSR.}

To address the inherent complexity imbalance between solvent and solute compounds, our framework, unsupervised hierarchical symbolic regression (UHiSR), is strategically organized hierarchically in a top-to-down manner. It is motivated by the rationale grounded in established practices observed in TLC experiments. As illustrated in Fig.~\ref{fig:learn_variable}, each stage focuses on specific aspects of the feature space, allowing the model to learn different latent variables which discern patterns at different levels of granularity. 

For instance, the top stage provides a broad overview of the interaction between solvent and solute, capturing their global polarity characteristics. Subsequent stages then delve into more detailed and specific characteristics of the solute molecular structure, offering intricate insights regarding the contributions of different molecular features to the overall molecular polarity.
As illustrated in Fig.~\ref{Fig.1}C, the framework contains three key parts: feature engineering, hierarchical neural network and symbolic regressions. In this subsection, we introduce the algorithms and methods.

\paragraph*{Feature engineering.}
Organic molecules are composed of various functional groups arranged along their molecular skeletons. The arrangement of these groups influences the polarity of a given molecule. Organic chemists typically assess the polarity of target molecules by considering the type, number, and distribution of functional groups. Here, we focused on $16$ representative functional groups (Fig.~\ref{Fig.1}C). Besides the type and number of functional groups, their spatial distribution is also crucial for molecular polarity. To account for this, we included additional descriptors such as the number of aromatic rings, dipole moments, and molecular skeleton descriptors (detailed in the Supplementary Materials).

In terms of features related to the solvent, we considered the volume ratios ($V\%$) of five different solvent compounds, \emph{i.e.}, \textit{n}-Hexane, Ethyl acetate, Dichloromethane, Diethyl ether and Methanol.

To evaluate the representation power of the selected features, we conducted a comparative analysis. This analysis employed the XGBoost model and examined three different sets of features.  Hereto, like in previous studies~\cite{XU20223202}, we primarily focus on solute molecular features. To maintain consistency, the features related to the eluent solvent were kept the same across all experiments. Feature group A consists of $19$-dimensional solute features (Table~\ref{tab:feature}). Feature group B encompasses $167$-dimensional features, containing $167$-dimensional MACCS keys for describing molecular structure. Feature group C contains $8$-dimensional features, focusing on conventional physicochemical descriptors. Specifically, these descriptors include molecular weight (MW), topological polar surface area (TPSA), the logarithm of $n$-octanol–water partition coefficient (LogP), hydrogen bond donor number (HBD), hydrogen bond acceptor number (HBA), molecular polarity index (MPI), polar surface area percentage (PSA) and dipole moment (DM).   

\paragraph*{Architecture design of hierarchical neural network.}The architecture of our network is motivated by the rationale grounded in established practices observed in TLC experiments. TLC experiments conventionally entail the execution of controlled experiments designed to meticulously dissect the influence of solute polarity and solvent polarity on $R_f$. In this context, we modified conventional Multi-Layer Perceptron (MLP) architecture to meet the specific demand of our task. Specifically, given an input sample $\mathbf{x} \in \mathbf{R}^d$, the initial step involves the classification of all features into $k$ clusters, denoted as $\mathbf{x}_1, \dots, \mathbf{x}_k$. Here, $\mathbf{x}_i\in\mathbf{R}^{d_i}$ and their dimensions sum up to $d$, \emph{i.e.}, $\sum_{i=1}^k d_i=d$. To enhance interpretability of these clusters, we incorporated a chemists-guided strategy, and the clusters used at each stage are demonstrated in Fig.~\ref{fig:learn_variable}.

Then, a key innovation lies in the formulation of the target latent representation layer $j$,  where $z^{(j)}_i=g_i(\mathbf{x}_i)$. Each function $g_i$ operates independently in the modified MLP, in the sense that there is no parameter sharing across $g_1, \dots, g_k$. Thus, we can consider $g_i$ as a sub-model. This design ensures the preservation of feature-specific information. The final output $o=h(\mathbf{z}^{(j)})$ encapsulates the information from the target latent representation layer. 

This design restricts the receptive field of the latent variables, therefore ensuring that each latent variable solely depends on a specific group of input features.  For instance, in the first stage, we considered all features and categorized them into two distinct clusters, solvent features and solute features. This categorization aligns with practical considerations where the polarity of the solvent and the solute is often analyzed separately. Consequently, the learned latent variables in this layer can be regarded as the solvent polarity index and the solute polarity index, respectively. The connections between the latent variables and the output layer capture the nonlinear relationship inherent in the data. This approach enables the model to learn informative latent variables in an unsupervised manner, which disentangles and highlights the distinct contributions of different groups of input features.

\paragraph*{Symbolic regression.}SR aims to identify mathematical expressions that capture the inherent relationships within a given dataset. Unlike approaches that involve fitting parameters to overly complex general models (such as machine learning models), SR explores a space of simple analytic expressions. The goal is to discover accurate and interpretable models that directly represent the underlying patterns in the data. SR is usually implemented by evolutionary algorithms, such as genetic programming (GP)~\cite{billard2003statistics,arnaldo2014multiple,quade2016prediction}. 

The most common way to visualize a symbolic expression is a tree-structure with nodes and branches, as shown in Fig.~\ref{fig:sigmoid}A. This structure includes primitive functions (\emph{e.g.}, $+, -, \times, \div, \exp)$ and terminal nodes (input features and numeric constants). Through a series of mutation operations, the GP algorithm seeks to determine the optimal number of nodes and terminals that provide the best fit to a given dataset.  For our implementation of the widely used GP-SR algorithm, we employed the open-source Python library PySR\footnote{\href{https://github.com/MilesCranmer/PySR}{https://github.com/MilesCranmer/PySR}} \cite{cranmer2023interpretable}. The hyperparameter configurations for our implementation are detailed in the Supplementary Materials. 

It is worth mentioning that our method is not limited to GP algorithms. In fact, the UHiSR framework is designed to accommodate various symbolic regression methods; comparable results obtained from deep reinforcement learning based SR methods (such as DISCOVER \cite{du2022discover}) are included in the Supplementary Materials. 

\subsection*{Experimental settings and training details}
When constructing the hierarchical neural network, at each stage we set the number of neurons in the hidden layers of each sub-model to $50$. Furthermore, two hidden layers were employed. To facilitate the discovery of concise equations, we leaned towards employing a relatively simple network structure, which helps to prevent overfitting to experimental noise. The activation function chosen for each sub-model was LeakyReLU. In the first stage, the activation function of the output layer was set to a sigmoid, which ensures a standardized output within the range of $0$ and $1$ (Fig.~\ref{fig:sigmoid}B). The batch size was set to $2048$, and the Adam optimizer implemented in PyTorch was applied. The learning rate was $0.01$. The dataset was randomly divided into $80/10/10$ for train/valid/test splits. Each hierarchical neural network was trained for $1000$ epochs, then the best validation checkpoint was selected for testing. To assess the prediction performance, we employed two key metrics: the $R$-squared coefficient ($R^2$) and the root mean square error (RMSE), which are defined as
\begin{equation}
    R^2 = 1-\frac{\sum_{i=1}^N(y_i-\hat{y}_i)^2}{\sum_{i=1}^N(y_i-\bar{y})^2}, \quad {\rm RMSE} = \sqrt{\frac{\sum_{i=1}^N(y_i-\hat{y}_i)^2}{N}},
\end{equation}
where $N$ is the total number of samples, $y_i$ and $\hat{y}_i$ represent the label and the prediction value for the $i$-th sample, and $\bar{y}$ denotes the mean of the ground truth values.

% Your references go at the end of the main text, and before the
% figures.  For this document we've used BibTeX, the .bib file
% scibib.bib, and the .bst file Science.bst.  The package scicite.sty
% was included to format the reference numbers according to *Science*
% style.

%BibTeX users: After compilation, comment out the following two lines and paste in
% the generated .bbl file. 

\bibliography{scibib}
\bibliographystyle{Science}

\section*{Acknowledgments}
We thank the High Performance Computing Platform of Peking University for machine learning model training. We also thank Mengge Du for assisting in the implementation of DISCOVER.

\paragraph*{Funding:} 
National Natural Science Foundation of China No.22071004 \\
National Natural Science Foundation of China No.21933001 \\
National Natural Science Foundation of China No.22150013 \\
National Natural Science Foundation of China No.62106116  \\

\paragraph*{Author contributions:}
Conceptualization: YC, FM \\
Methodology: SL, CL, YC, FM \\
Investigation: SL, CL \\
Visualization: SL, CL \\
Supervision: YC, FM \\
Writing-original draft: SL, CL  \\
Writing-review \& editing: SL, YC, FM  \\

\paragraph*{Competing interests:} Authors declare that they have no competing interests.

\paragraph*{Data and materials availability:}
The dataset utilized in this manuscript is accessible at the following link: \href{https://github.com/SiyuLou/UnsupervisedHierarchicalSymbolicRegression}{https://github.com/SiyuLou/UnsupervisedHierarchicalSymbolicRegression}. \\
All original code for the experiments and analysis in this work has been deposited at the website \href{https://github.com/SiyuLou/UnsupervisedHierarchicalSymbolicRegression}{https://github.com/SiyuLou/UnsupervisedHierarchicalSymbolicRegression}.
%Here you should list the contents of your Supplementary Materials -- below is an example. 
%You should include a list of Supplementary figures, Tables, and any references that appear only in the SM. 
%Note that the reference numbering continues from the main text to the SM.
% In the example below, Refs. 4-10 were cited only in the SM.     

\newpage
\section*{Supplementary Materials}
\setcounter{table}{0}
\setcounter{figure}{0}
\renewcommand{\thetable}{S\arabic{table}}
\renewcommand\thefigure{S\arabic{figure}}

\subsection*{Candidate equations for the hierarchical equation system}
We present the candidate equations for each the hierarchical equation system, along with the corresponding $R^2$ and RMSE values on the test dataset as reported in the Table S1-S10. Symbolic regression algorithms often face a trade-off between simplicity and fitting performance. Considering the noise in real-world datasets, we prioritize simpler equations, such as choosing $R_f= \sigma(3.48 \Psi + 3.08 \xi + 1.86)$ as the $R_f$ governing equation, even if there is a slight loss in fitting accuracy. On one hand, concise equations are easier to understand; on the other hand, they help avoid overfitting to noise, making the model more robust.

\begin{table}[h]
\centering
\caption{Candidate equations for $R_f \sim (\Psi, \xi)$.}
\begin{tabular}{@{}lccc@{}}
\hline
Equation & $R^2$ &RMSE \\
\hline
{\footnotesize $R_f= \sigma(3.48 \Psi + 3.08 \xi + 1.86),\quad \sigma(x) = 1/(1+e^{-x})$ } & $0.930$ &$0.091$ \\[0.2cm]
{\footnotesize $R_f= \sigma(3.48 \Psi + 0.696 \xi^{2} + 3.48 \xi + 1.86),\quad \sigma(x) = 1/(1+e^{-x})
$} & $0.931$ & $0.090$ \\[0.2cm]
\begin{minipage}{0.8\linewidth}  {\footnotesize $ R_f= \sigma(3.48 \Psi + \frac{\xi^{4}}{ \Psi^{2} + 1.44\xi^{2}} + 3.48 \xi + 1.86),\quad \sigma(x) = 1/(1+e^{-x})$ } \end{minipage} & $0.933$ & $0.088$ \\[0.2cm]
\begin{minipage}{0.8\linewidth}  {\footnotesize $ R_f= \sigma(3.48 \Psi + \frac{\xi^{4}}{8.40 \Psi^{4} + 1.44\xi^{2}} + 3.48 \xi + 1.86),\quad \sigma(x) = 1/(1+e^{-x})
 $} \end{minipage}  & $0.934$ & $0.088$ \\[0.2cm]
\begin{minipage}{0.8\linewidth}  {\footnotesize $ R_f=\sigma(3.42 \Psi + 3.48 \xi + 1.86 + \frac{2 \Psi^{3} \xi^{2} + \Psi \xi^{2}  + 1.01 \xi^{4}}{6.33 \Psi^{4} + 1.44 \xi^{2} }),\quad \sigma(x) = 1/(1+e^{-x})$ } \end{minipage}& $0.934$ & $0.088$ \\[0.2cm]
\hline
\end{tabular}
\end{table}

\begin{table}[h]
\centering
\caption{Candidate equations for $\Psi \sim ($Hex, EA, DCM, MeOH, Et$_2$O$)$.}
\begin{tabular}{@{}lccc@{}}
\hline
Equation & $R^2$ & RMSE\\
\hline
{\footnotesize $\Psi= - {\rm Hex} + 2 {\rm EA} - 0.259 {\rm DCM} + 12.5 {\rm MeOH} + {\rm Et}_2{\rm O}$} &$0.960$ &$0.144$\\[0.2cm]
{\footnotesize $\Psi=- {\rm Hex} + 1.59 {\rm EA} - 0.411 {\rm DCM} + 11.1 {\rm MeOH} + {\rm Et}_2{\rm O}^{2} + 0.142$}& $0.983$ & $0.094$ \\[0.2cm]
\begin{minipage}{0.8\linewidth}  {\footnotesize $ \Psi= - {\rm Hex}  + 1.59  {\rm EA}  + \frac{1.55 {\rm DCM}^{2} {\rm MeOH}}{0.102 -  {\rm EA} } - 0.411 {\rm DCM} + {\rm Et}_2{\rm O}^{2} + 0.141
$} \end{minipage} & $0.987$ & $0.082$ \\[0.2cm]
\begin{minipage}{0.8\linewidth}  {\footnotesize $ \Psi=- {\rm Hex}  + 1.56  {\rm EA}  + \frac{1.31 {\rm DCM}^{2} {\rm MeOH}}{{\rm Et}_2{\rm O} + 0.0874} - 0.436 {\rm DCM} + {\rm Et}_2{\rm O}^{2} + 0.145
 $} \end{minipage}  & $0.988$ & $0.079$ \\[0.2cm]
\begin{minipage}{0.8\linewidth}  {\footnotesize $ \Psi=- {\rm Hex} + 1.52  {\rm EA}  + 21.1 {\rm DCM}^{3} {\rm MeOH} \left({\rm Hex}  + {\rm DCM}\right)^{2} - 0.477 {\rm DCM} - {\rm MeOH} $\\$
~\quad\quad+ {\rm Et}_2{\rm O}^{2} + 0.149$} \end{minipage}& $0.989$ & $0.074$ \\[0.2cm]
\hline
\end{tabular}
\end{table}

\begin{table}
\centering
\caption{Candidate equations for $\xi \sim (\alpha, \beta)$.}
\begin{tabular}{@{}lccc@{}}
\hline
Equation & $R^2$ &RMSE \\
\hline
{\footnotesize $\xi=0.367 + \frac{\beta + 1.89}{\alpha - 1.47} $}& $0.827$ &$0.28$ \\[0.2cm]
{\footnotesize $\xi = - 0.232 \beta - 0.232 \left(e^{\alpha} - 0.0531\right) \left(- 2 \alpha + 2 \beta + 4.71\right)$} & $0.841$ & $0.270$ \\[0.2cm]
\begin{minipage}{0.8\linewidth}  {\footnotesize $ \xi =- 0.232 \beta - 0.232 \left(e^{\alpha} - 0.0531\right) \left(- 3 \alpha + \frac{\alpha}{4 \beta^{2}} + 2 \beta + 4.71\right) $} \end{minipage} & $0.858$ & $0.255$ \\[0.2cm]
\begin{minipage}{0.8\linewidth}  {\footnotesize $ \xi = - 0.232 \beta - 0.232 \left(e^{\alpha} - 0.0531\right) \left(- 3 \alpha + 2 \beta + \left(4.70 \alpha + 4.70 \beta\right) e^{- \beta^{2}} + 4.71\right)
 $} \end{minipage}  & $0.875$ & $0.238$ \\[0.2cm]
\begin{minipage}{0.8\linewidth}  {\footnotesize $ \xi = - 0.232 \beta - 0.232 \left(e^{\alpha} \!-\! 0.0531\right) \left(- 3 \alpha + 2 \beta + \left(4.72 \alpha + 4.72 \beta\right) e^{- \left(\beta - 0.160\right)^{2}}\! + \!4.71\right)$} \end{minipage}& $0.878$ & $0.236$ \\[0.2cm]
\hline
\end{tabular}
\end{table}

\begin{table}
\centering
\caption{Candidate equations for $\alpha \sim ($ NBen, MSD, DM $)$.}
\begin{tabular}{@{}lccc@{}}
\hline
Equation & $R^2$ &RMSE \\
\hline
{\footnotesize $\alpha=\frac{- 2 {\rm NBen} - \frac{1.33 {\rm NBen}}{{\rm MSD} - 0.0467}}{{\rm DM} + 0.412} - 0.743 $}& $0.898$ &$0.63$ \\[0.2cm]
{\footnotesize $\alpha =\frac{- 2 {\rm NBen} - \frac{1.82 \left({\rm NBen} + {\rm MSD}\right)}{{\rm MSD}^{2} + 1.12}}{{\rm DM} + 0.405} - 0.529$} & $0.907$ & $0.604$ \\[0.2cm]
\begin{minipage}{0.8\linewidth}  {\footnotesize $\alpha = \frac{- 2 {\rm NBen} + \frac{\left({\rm NBen} + {\rm MSD}\right) \left(0.147 {\rm MSD} - 1.10\right)}{{\rm MSD} + 0.0978}}{{\rm DM} + 0.405} - 0.512$} \end{minipage} & $0.912$ & $0.587$ \\[0.2cm]
\begin{minipage}{0.8\linewidth}  {\footnotesize $ \alpha = \frac{- 2 {\rm NBen} + \frac{\left({\rm NBen} + {\rm MSD}\right) \left(0.147 {\rm MSD} - 1.10\right)}{{\rm MSD} + 0.0978}}{{\rm DM} + 0.405} - 0.512
 $} \end{minipage}  & $0.922$ & $0.552$ \\[0.2cm]
\begin{minipage}{0.8\linewidth}  {\footnotesize $ \alpha=\frac{- 2 {\rm NBen} + \frac{\left({\rm NBen}^{2} + \left({\rm DM} + 0.386\right) \left(- {\rm NBen}^{2} + {\rm NBen} + {\rm MSD}\right)\right) \left(0.153 {\rm MSD} + 0.153 {\rm DM} - 1.27\right)}{{\rm MSD} + 0.327}}{{\rm DM} + 0.405} - 0.529 $} \end{minipage}& $0.923$ & $0.549$ \\[0.2cm]
\hline
\end{tabular}
\end{table}

\begin{table}
\centering
\caption{Candidate equations for $\beta \sim (\gamma_1, \gamma_2, \gamma_3, \gamma_4, \gamma_5)$.}
\begin{tabular}{@{}lccc@{}}
\hline
Equation & $R^2$ &RMSE \\
\hline
{\footnotesize $\beta = \left(\gamma_1 + \gamma_5\right) \left(0.0212 \gamma_3 \gamma_4 - 0.441\right) + e^{0.238 \gamma_2 + 0.238 \gamma_4} - 0.488$} & $0.774$ &  $1.02$ \\[0.2cm]
{\footnotesize $\beta=- \frac{0.218 \gamma_1}{\gamma_2^{6}} + 0.412 \gamma_2 + 0.412 \gamma_4 - 0.412 \gamma_5 + 0.0212 \left(\gamma_3 + \gamma_4\right)^{2} + 0.333$}& $0.804$ &$0.950$ \\[0.2cm]
{\footnotesize $\beta =- \frac{0.218 \gamma_1}{\gamma_2^{6}} + 0.413 \gamma_2 + 0.435 \gamma_4 - 0.435 \gamma_5 + 0.0223 \left(\gamma_3 + \gamma_4\right)^{2} + 0.493 $} & $0.810$ & $0.934$ \\[0.2cm]
\begin{minipage}{0.8\linewidth}  {\footnotesize $\beta = - \frac{0.241 \gamma_1}{\gamma_2^{6}}  + 0.367 $\\$~\quad\quad+ 0.241 \left(- 0.0933 \gamma_1 - 0.0933 \gamma_3 - 1.52\right) \left(- \gamma_2 - \gamma_4 + \gamma_5 - 0.0820 \left(\gamma_3 + \gamma_4\right)^{2}\right) $} \end{minipage} & $0.823$ & $0.904$ \\[0.2cm]
\begin{minipage}{0.8\linewidth}  {\footnotesize $ \beta =  - \frac{0.241 \gamma_1}{\gamma_2^{6}} + 0.367 $\\$~\quad\quad+0.241 \left(- 0.150 \gamma_1 - 0.150 \gamma_3 - 1.52\right)\left(- \gamma_2 - \gamma_4 + \gamma_5 - 0.0784 \left(\gamma_3 + \gamma_4\right)^{2} - 0.501\right) 
 $} \end{minipage}  & $0.831$ & $0.883$ \\[0.2cm]
\hline
\end{tabular}
\end{table}

\begin{table}
\centering
\caption{Candidate equations for $\gamma_1 \sim ($CtAmides, CtCO$_2$H$)$.}
\begin{tabular}{@{}lccc@{}}
\hline
Equation & $R^2$ &RMSE \\
\hline
{\footnotesize $\gamma_1 =- 3.26 {\rm CtAmides} - 3.26 {\rm CtCO}_2{\rm H} + 1.87$} & $0.989$ &  $0.152$ \\[0.2cm]
{\footnotesize $\gamma_1= - 3.09 {\rm CtAmides} - 3.91 {\rm CtCO}_2{\rm H} + 1.87 $}& $0.997$ &$0.074$ \\[0.2cm]
{\footnotesize $\gamma_1 = - 2 {\rm CtAmides} - {\rm CtCO}_2{\rm H} \left({\rm CtAmides} + {\rm CtCO}_2{\rm H} + 2.66\right) + 1.86 e^{- {\rm CtAmides}^{2}}$} & $0.999$ & $0.032$ \\[0.2cm]
\begin{minipage}{0.7\linewidth}  {\footnotesize $\gamma_1= - 2 {\rm CtAmides} - {\rm CtCO}_2{\rm H} \left({\rm CtAmides} + {\rm CtCO}_2{\rm H} + 1.57\right) $\\$~\quad ~\quad+ 1.86 e^{- {\rm CtAmides} - {\rm CtCO}_2{\rm H}}$} \end{minipage} & $0.999$ & $0.035$ \\[0.3cm]
\begin{minipage}{0.7\linewidth}  {\footnotesize $ \gamma_1= - 2 {\rm CtAmides} - 0.617 {\rm CtCO}_2{\rm H}^{2} - 3.15 {\rm CtCO}_2{\rm H} $\\$~\quad~\quad+ 1.87 e^{- {\rm CtAmides} \left({\rm CtAmides} + {\rm CtCO}_2{\rm H}\right)} $} \end{minipage}  & $1.000$ & $0.012$ \\[0.2cm]
\hline
\end{tabular}
\end{table}

\begin{table}
\centering
\caption{Candidate equations for $\gamma_2 \sim ($CtRNH$_2$, CtOH, CtPhenol$)$.}. 
\begin{tabular}{@{}lccc@{}}
\hline
Equation & $R^2$ &RMSE \\
\hline
{\footnotesize $\gamma_2 =1.86 \left(- 0.734 {\rm CtRNH}_2^{2} + 0.734 {\rm CtOH} - 0.734 {\rm CtPhenol} + 1\right)^{2} - e^{{\rm CtPhenol}}$} & $0.955$ &  $0.389$ \\[0.2cm]
\begin{minipage}{0.8\linewidth} {\footnotesize $\gamma_2= {\rm CtRNH}_2^{2} - {\rm CtPhenol} - 2.79$\\$~\quad~\quad + 3.59 \left(- 0.528 {\rm CtRNH}_2^{2} + 0.528 {\rm CtOH} - 0.528 {\rm CtPhenol} + 1\right)^{2} $}\end{minipage}& $0.987$ &$0.204$ \\[0.2cm]
\begin{minipage}{0.8\linewidth} {\footnotesize $\gamma_2 = - {\rm CtRNH}_2 + 2 {\rm CtOH} - 1.76 {\rm CtPhenol} \cdot \left(1.76 - {\rm CtRNH}_2\right) $\\$~\quad~\quad+ \left(- {\rm CtRNH}_2^{2} + {\rm CtOH} - {\rm CtPhenol} + 0.912\right)^{2}$ }\end{minipage}& $0.999$ & $0.071$ \\[0.2cm]
\begin{minipage}{0.8\linewidth}  {\footnotesize$\gamma_2=- 1.06 {\rm CtRNH}_2 + 2 {\rm CtOH} - {\rm CtPhenol} \cdot \left(2.13 - 2 {\rm CtRNH}_2\right) - {\rm CtPhenol} $\\$~\quad~\quad+ \left(- {\rm CtRNH}_2^{2} + {\rm CtOH} - {\rm CtPhenol} + 0.912\right)^{2}$} \end{minipage} & $0.999$ & $0.060$ \\[0.2cm]
\begin{minipage}{0.8\linewidth}  {\footnotesize $ \gamma_2= - 1.13 {\rm CtRNH}_2 \left(1 - 0.939 {\rm CtOH}\right)^{2} + {\rm CtOH}^{3} + {\rm CtOH} $\\$~\quad~\quad- {\rm CtPhenol} \left(- {\rm CtRNH}_2^{2} - 0.853 {\rm CtRNH}_2 + 2.20\right) - {\rm CtPhenol} $\\[0.2cm]$~\quad~\quad+ \left(- {\rm CtRNH}_2^{2} + {\rm CtOH} - {\rm CtPhenol} + 0.912\right)^{2} $} \end{minipage}  & $1.000$ & $0.036$ \\[0.2cm]
\hline
\end{tabular}
\end{table}

\begin{table}
\centering
\caption{Candidate equations for $\gamma_3 \sim ($CtNO$_2$, CtRCO$_2$R$)$.}
\begin{tabular}{@{}lccc@{}}
\hline
Equation & $R^2$ &RMSE \\
\hline
{\footnotesize $\gamma_3 ={\rm CtNO}_2 \left(- {\rm CtRCO}_2{\rm R}^{2} - 3.65\right) + 0.762$} & $0.961$ &  $0.413$ \\[0.2cm]
\begin{minipage}{0.8\linewidth} {\footnotesize $\gamma_3= {\rm CtNO}_2 \left(- {\rm CtRCO}_2{\rm R}^{2} - 3.65\right) + 0.872 e^{- 8 {\rm CtRCO}_2{\rm R}^{3}}$}\end{minipage}& $0.979$ &$0.305$ \\[0.2cm]
\begin{minipage}{0.8\linewidth} {\footnotesize $\gamma_3 ={\rm CtNO}_2 \left(- {\rm CtRCO}_2{\rm R}^{2} - 3.65\right) - \frac{{\rm CtRCO}_2{\rm R}}{{\rm CtNO}_2 + {\rm CtRCO}_2{\rm R}^{3} - 0.329} + 0.853$} \end{minipage}& $0.991$ & $0.202$ \\[0.2cm]
\begin{minipage}{0.8\linewidth}  {\footnotesize $\gamma_3=- 0.346 {\rm CtNO}_2^{2} {\rm CtRCO}_2{\rm R}^{2} - {\rm CtRCO}_2{\rm R} $\\$~\quad~\quad+ \left({\rm CtNO}_2 - 0.246\right) \left(- \frac{{\rm CtRCO}_2{\rm R}^{2}}{{\rm CtNO}_2^{3} + {\rm CtNO}_2 + {\rm CtRCO}_2{\rm R} - 1.52} - 3.70\right)$} \end{minipage} & $0.995$ & $0.150$ \\[0.2cm]
\begin{minipage}{0.8\linewidth}  {\footnotesize $ \gamma_3= - {\rm CtRCO}_2{\rm R} - 3.70\left({\rm CtNO}_2 - 0.246\right) $\\$~\quad~\quad+ \left({\rm CtNO}_2 \!-\! 0.246\right) \left(- 0.168 {\rm CtNO}_2^{7} {\rm CtRCO}_2{\rm R}^{3} - \frac{1.26 {\rm CtRCO}_2{\rm R}^{2}}{{\rm CtNO}_2^{3} + {\rm CtNO}_2 + {\rm CtRCO}_2{\rm R} - 1.52}\right) $} \end{minipage}  & $0.997$ & $0.116$ \\[0.2cm]
\hline
\end{tabular}
\end{table}

\begin{table}
\centering
\caption{Candidate equations for $\gamma_4 \sim ($CtF, CtAldehyde, CtR$_2$CO, CtCN, CtROR$)$.}
\begin{tabular}{@{}lccc@{}}
\hline
Equation & $R^2$ &RMSE \\
\hline
\begin{minipage}{0.8\linewidth} {\footnotesize $\gamma_4 = \left(- {\rm CtF} + \left(1.66 - {\rm CtAldehyde}\right) \left(- {\rm CtAldehyde} + {\rm CtR}_2{\rm CO} + {\rm CtCN}\right)\right) $\\$~\quad~\quad\left({\rm CtAldehyde}^{2} - {\rm CtROR}^{3} + 2.05\right) $} \end{minipage}& $0.937$ &  $0.565$ \\[0.2cm]
\begin{minipage}{0.8\linewidth} {\footnotesize $\gamma_4= - {\rm CtAldehyde} + \left(- {\rm CtAldehyde}^{2} - 4 {\rm CtROR}^{2} + 2.05\right) $\\$~\quad~\quad\left(- {\rm CtAldehyde} + 1.60 {\rm CtR}_2{\rm CO} + 1.60 {\rm CtCN} - {\rm CtF} + 0.0807\right)$}\end{minipage}& $0.939$ &$0.557$ \\[0.2cm]
\begin{minipage}{0.8\linewidth} {\footnotesize $\gamma_4 ={\rm CtAldehyde}^{3} - 3 {\rm CtAldehyde} + 3 {\rm CtR}_2{\rm CO}+ 2 {\rm CtCN} - 2 {\rm CtF} $\\$~\quad~\quad+ e^{{\rm CtCN} - \left({\rm CtROR} - 2 {\rm CtF}\right)^{2}} - 0.746$} \end{minipage}& $0.958$ & $0.459$ \\[0.2cm]
\begin{minipage}{0.8\linewidth}  {\footnotesize $\gamma_4 = {\rm CtAldehyde}^{3} - {\rm CtAldehyde} {\rm CtR}_2{\rm CO} - 3 {\rm CtAldehyde} + {\rm CtR}_2{\rm CO}^{2} + 2 {\rm CtR}_2{\rm CO} $\\$~\quad~\quad + {\rm CtCN} - 2 {\rm CtF} + e^{{\rm CtCN} - \left({\rm CtROR} - 2 {\rm CtF}\right)^{2}} - 0.746$} \end{minipage} & $0.961$ & $0.446$ \\[0.2cm]
\begin{minipage}{0.8\linewidth}  {\footnotesize $ \gamma_4= {\rm CtAldehyde}^{3} - {\rm CtAldehyde}^{2} \left({\rm CtR}_2{\rm CO} - {\rm CtF}\right)^{2} - 3 {\rm CtAldehyde} + {\rm CtR}_2{\rm CO}^{2} $\\$~\quad~\quad+ 2 {\rm CtR}_2{\rm CO} + 4 {\rm CtCN} - 2 {\rm CtF} + e^{{\rm CtF} - \left({\rm CtROR} - 2 {\rm CtF}\right)^{2}} - 0.746 $} \end{minipage}  & $0.970$ & $0.391$ \\[0.2cm]
\hline
\end{tabular}
\end{table}

\begin{table}
\centering
\caption{Candidate equations $\gamma_5 \sim ($CtCl, CtBr, CtI, CtMethyl$)$.}
\begin{tabular}{@{}lccc@{}}
\hline
Equation & $R^2$ &RMSE \\
\hline
\begin{minipage}{0.8\linewidth} {\footnotesize $\gamma_5 =  \left({\rm CtCl} + 2 {\rm CtI}\right)^{2} + \left(\frac{{\rm CtBr}}{{\rm CtBr} - 0.305} + {\rm CtMethyl}\right)^{2}$} \end{minipage}& $0.931$ &  $0.805$ \\[0.2cm]
\begin{minipage}{0.8\linewidth} {\footnotesize $\gamma_5= {\rm CtI} {\rm CtMethyl} + \left({\rm CtCl} + 2 {\rm CtI}\right)^{2} + \left(\frac{{\rm CtBr}}{{\rm CtBr} - 0.359} + {\rm CtMethyl}\right)^{2} - 0.372$}\end{minipage}& $0.951$ &$0.677$ \\[0.2cm]
\begin{minipage}{0.8\linewidth} {\footnotesize $\gamma_5 ={\rm CtBr} - \left({\rm CtBr}^{3} - {\rm CtMethyl}\right)^{2} \left({\rm CtCl} + {\rm CtI} + 0.245\right)^{2} $\\$~\quad~\quad+ \left({\rm CtCl} + {\rm CtBr} + 2 {\rm CtI} + {\rm CtMethyl}\right)^{2} - 0.330$} \end{minipage}& $0.965$ & $0.578$ \\[0.2cm]
\begin{minipage}{0.8\linewidth}  {\footnotesize $\gamma_5 =- {\rm CtCl}^{2} + 2 {\rm CtCl} + {\rm CtBr}- \left({\rm CtCl} + {\rm CtI}\right)^{3} \left({\rm CtBr} + {\rm CtMethyl}^{2} + {\rm CtMethyl}\right) $\\$~\quad~\quad- 0.486 \left({\rm CtBr}^{2} - 0.879\right)^{2} + \left({\rm CtCl} + {\rm CtBr} + 2 {\rm CtI} + {\rm CtMethyl}\right)^{2}$} \end{minipage} & $0.982$ & $0.411$ \\[0.2cm]
\begin{minipage}{0.8\linewidth}  {\footnotesize $ \gamma_5=  - {\rm CtCl}^{2} + {\rm CtCl} {\rm CtI} + 1.46 {\rm CtCl} + {\rm CtBr} $\\$~\quad~\quad - \left({\rm CtCl} + {\rm CtI}\right)^{3} \left({\rm CtBr} + {\rm CtMethyl}^{2} + {\rm CtMethyl}\right) $\\$~\quad~\quad- 0.486 \left({\rm CtBr}^{2} - 0.879\right)^{2} + \left({\rm CtCl} + {\rm CtBr} + 2 {\rm CtI} + {\rm CtMethyl}\right)^{2} $} \end{minipage}  & $0.985$ & $0.377$ \\[0.2cm]
\hline
\end{tabular}
\end{table}

\subsection*{Molecular skeleton descriptor}
\begin{figure}[h]
\centering
  \includegraphics[width=\textwidth]{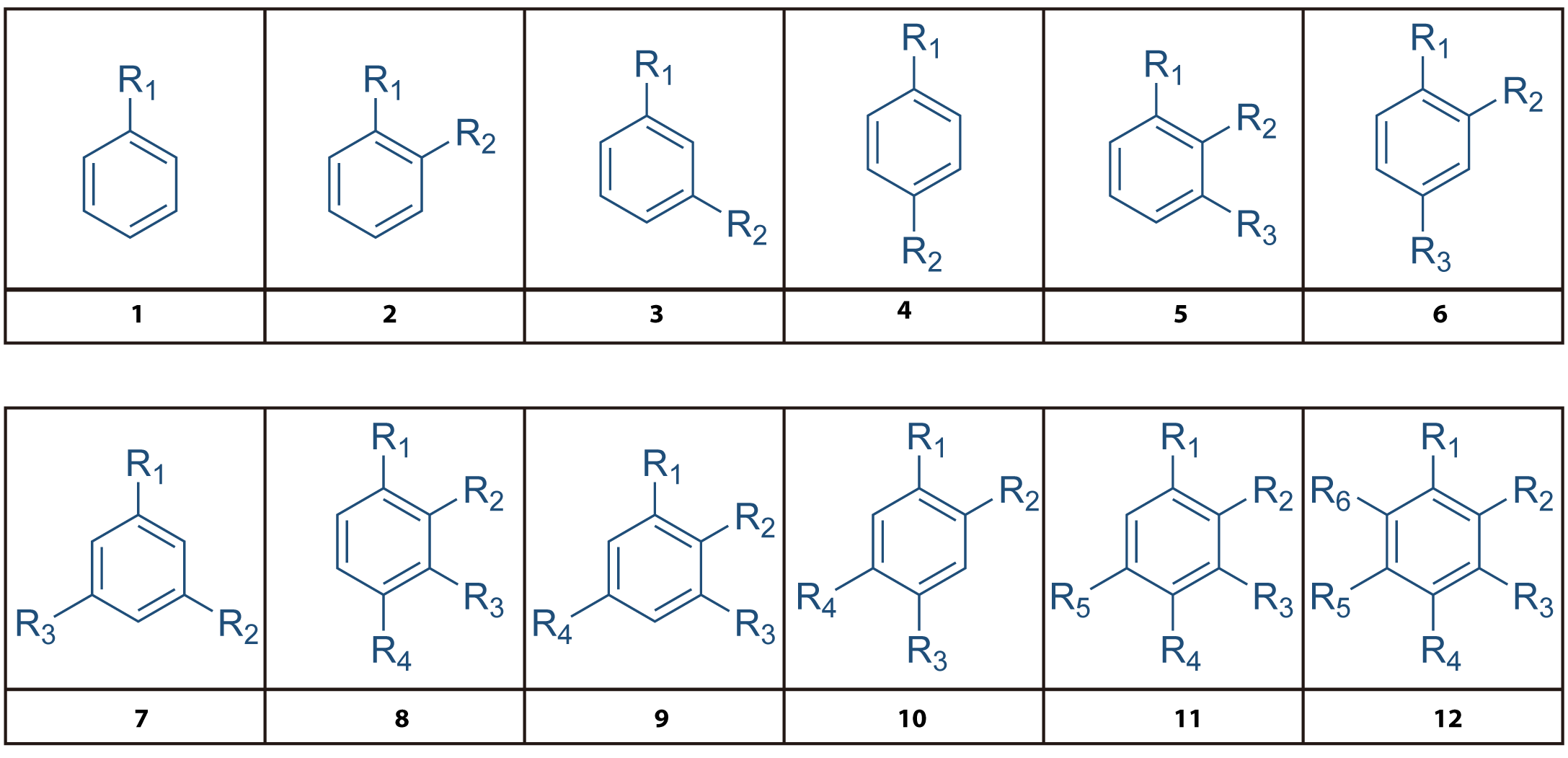}
  \caption{\textbf{Illustration of molecular skeleton descriptors.}}
  \label{figs1}
\end{figure}

The distribution of functional groups within a molecule influences its properties. To quantify this effect, we developed a molecular skeleton descriptor (MSD) focusing on aromatic rings. In the MSD framework, the benzene ring constitutes the core structure. The descriptor values are based on the pattern of substitution on the benzene ring. The value $0$ indicates the absence of a benzene ring. The value $1$ denotes a mono-substituted aromatic ring. The values $2$, $3$, and $4$ represent di-substituted aromatic rings in ortho, meta, and para positions, respectively. Values $5$, $6$, and $7$ correspond to tri-substituted aromatic rings with $1$,$2$,$3$-substitution, $1$,$2$,$4$-substitution, and $1$,$3$,$5$-substitution patterns, respectively. Furthermore, values $8$, $9$, and $10$ are assigned to tetra-substituted aromatic rings with $1$,$2$,$3$,$4$-substitution, $1$,$2$,$3$,$5$-substitution, and $1$,$2$,$4$,$5$-substitution patterns, respectively. Lastly, values $11$ and $12$ are used to denote penta-substituted and hexa-substituted aromatic rings, respectively. $12$ MSD structures with the aromatic ring are illustrated in Fig.~\ref{figs1}. This descriptor provides a systematic approach to categorize the substitution patterns of benzene rings within molecular structures.

\subsection*{Hyperparameter configurations for pySR}
We implemented the GP algorithm using the open-source Python library pySR. The hyperparameter configurations are detailed in Tables~\ref{tab:pySR1} and \ref{tab:pySR2}. To capture the higher non-linearity in the relationships between various polarity indices (\emph{e.g.}, $\alpha$, $\beta$) and their corresponding input features, we increased the number of cycles per iteration. This adjustment aims to discover equations that have better fitting performance, with a trade-off of increased computation time.
\begin{table}[h]
\caption{Hyperparameters setup for discovering $R_f$ governing equation.}
\centering
\footnotesize % This will make the font size of the entire table smaller
\begin{tabular}{@{}lc@{}}
\hline
Hyperparameter & Value  \\
\hline
procs & $4$ \\
population & $8$ \\
population size & $50$ \\
ncyclesperiteration & $50$ \\
niterations & $200$ \\
maxsize & $50$ \\
maxdepth & $10$ \\
binary operators & [``$+$'', ``$\times$'', ``$-$'', ``$/$'']  \\
unary operators & ``square'' \\
complexity of constants & $2$ \\
\hline
\end{tabular}
\label{tab:pySR1}
\end{table}

\begin{table}[h]
\caption{Hyperparameters setup for discovering polarity indices governing equation.}
\centering
\footnotesize % This will make the font size of the entire table smaller
\begin{tabular}{@{}lc@{}}
\hline
Hyperparameter & Value  \\
\hline
procs & $4$ \\
population & $8$ \\
population size & $50$ \\
ncyclesperiteration & $500$ \\
niterations & $200$ \\
maxsize & $50$ \\
maxdepth & $10$ \\
binary operators & [``$+$'', ``$\times$'', ``$-$'', ``$/$''] \\
unary operators & [``square'', ``cube'', ``exp''] \\
complexity of constants & $2$ \\
\hline
\end{tabular}
\label{tab:pySR2}
\end{table}

\subsection*{More results for various SR algorithms}
The UHiSR framework is flexible and supports multiple symbolic regression methods. In addition to the GP algorithm, we also present results from SR algorithms employing deep reinforcement learning. The open source Python package, DISCOVER\footnote{\href{https://github.com/menggedu/discover}{https://github.com/menggedu/discover}}, was utilized for the implementation of this method. The discovered hierarchical equation system is detailed in Table~\ref{tab:discover}. The hyperparameter configurations for our implementations are detailed in Table~\ref{tab:discover-hyper}.

\begin{table}[h]
\centering
\caption{Hierarchical equation systems from DISCOVER.}
\begin{tabular}{@{}lcc@{}}
\hline
Equation & $R^2$ &RMSE \\
\hline
{\footnotesize $R_f = \sigma\left(3.48 \Psi + 3.08 \xi + 1.86\right),\quad \sigma(x) = 1/(1+e^{-x})$ } & $0.930$ &$0.091$\\[0.2cm]
{\footnotesize $\Psi=0.73 {\rm EA} + 7.37{\rm MeOH} -0.27 {\rm DCM} + 0.51 {\rm Et}_2{\rm O}^2 -1.29{\rm Hex}^4$} & $0.962$ & $0.11$ \\[0.2cm]
\begin{minipage}{0.8\linewidth}  {\footnotesize $ \xi =- 0.03\alpha + 0.04(e^\alpha - \beta)e^\alpha - 0.21\beta-1.57e^\alpha $} \end{minipage} & $0.840$ & $0.271$ \\[0.2cm]
\hspace{1.7em}{\footnotesize $\alpha = -1.97{\rm NBen} + 0.22{\rm DM} -1.49e^{{\rm NBen}-{\rm NBen}\times{\rm DM}}+0.12{\rm NBen}\times{\rm MSD}$} &$0.867$& $0.722$ \\[0.2cm]
\begin{minipage}{0.8\linewidth}\hspace{1.7em}{\footnotesize $\beta=0.45\gamma_4 + 0.026(\gamma_5-\gamma_1)^2 -0.33\gamma_1 + 0.064 \gamma_2^2 + 0.036\gamma_4^2 + 0.055 \gamma_3\gamma_4 -0.56 \gamma_5$\\$~\quad~\quad~\quad~+ 0.27 \gamma_2$} \end{minipage} & $0.793$&$0.975$\\[0.2cm]
\hspace{1.7em}{\footnotesize $\gamma_1=1.88e^{\rm CtAmides} -3.93{\rm CtAmides} -2.48 {\rm CtAmides}^2 -3.92{\rm CtCO}_2{\rm H}$}& $0.999$&$0.046$\\[0.2cm]
\begin{minipage}{0.8\linewidth}\hspace{1.7em}{\footnotesize $\gamma_2=-7.95{\rm CtRNH}_2 + 9.30{\rm CtOH} -5.98{\rm CtPhenol} + 1.09e^{{\rm CtPhenol}-{\rm CtRNH}_2}$\\$~\quad~\quad\quad\quad-2.66 e^{\rm CtOH} + 2.41 e^{{\rm CtRNH}_2}$}  \end{minipage}& $0.996$&$0.046$\\[0.2cm]
\begin{minipage}{0.8\linewidth}\hspace{1.7em}{\footnotesize $\gamma_3=0.93e^{{\rm CtCO}_2{\rm H} - {\rm CtAmides}} -3.14{\rm CtAmides}-0.014e^{{\rm CtAmides}}-2.81{\rm CtCO}_2{\rm H} $\\$~\quad~\quad\quad\quad-0.001 e^{{\rm CtCO}_2{\rm H}^2}$}
\end{minipage}& $0.991$&$0.203$\\[0.2cm]
\begin{minipage}{0.8\linewidth}\hspace{1.7em}{\footnotesize $\gamma_4 = 1.92 {\rm CtCN}-0.61{\rm CtR}_2{\rm CO} + 1.92{\rm CtCN} + 0.32e^{{\rm CtAldehyde}} + 2.34 {\rm CtAldehyde}$\\$~\quad~\quad\quad\quad -0.095 {\rm CtROR}^3 -4.32 {\rm CtROR} + 0.99{\rm CtROR}^2 + 0.51 {\rm CtF}^4 -3.00{\rm CtF}$} 
\end{minipage}& $0.971$&$0.385$\\[0.2cm]
\begin{minipage}{0.8\linewidth}\hspace{1.7em}{\footnotesize $\gamma_5 = 3.66{\rm CtBr} + 2.43{\rm CtMethyl} + 0.63{\rm CtCl} -0.64 e^{{\rm CtBr}} + 0.57({\rm CtCl}+4{\rm CtI})^2 $\\$~\quad~\quad\quad\quad-4.42{\rm CtI}$}    
\end{minipage} & $0.935$&$0.780$\\[0.2cm]
\hline
\end{tabular}
\label{tab:discover}
\end{table}

\begin{table}[h]
\caption{Hyperparameters setup for DISCOVER.}
\centering
\footnotesize % This will make the font size of the entire table smaller
\begin{tabular}{@{}llc@{}}
\hline
Category & Hyperparameter & Value  \\
\hline
task & function set & [``add'', ``sub'', ``mul'', ``div'',``n2'',``n3'',``exp''] \\
training & n samples & $200000$ \\
&batch size & $500$ \\
&epsilon & $0.2$ \\
controller & learning rate & $0.0025$ \\
& entropy weight & $0.03$ \\
& entropy gamma & $0.7$ \\

\hline
\end{tabular}
\label{tab:discover-hyper}
\end{table}
\end{document}